\documentclass[letterpaper]{article} 
\usepackage{aaai24}  
\usepackage{times}  
\usepackage{helvet}  
\usepackage{courier}  
\usepackage[hyphens]{url}  
\usepackage{graphicx} 
\usepackage{natbib}  
\usepackage{caption} 
\usepackage{algorithm}
\usepackage{algorithmic}
\usepackage{amsmath}
\usepackage{amssymb}
\usepackage{xspace}
\usepackage{colortbl}
\usepackage{array} 
\usepackage{xcolor}
\usepackage{makecell}
\usepackage{bm}
\usepackage{subfigure}
\usepackage{arydshln} 
\usepackage[inline]{enumitem}

\makeatletter
\DeclareRobustCommand\onedot{\futurelet\@let@token\@onedot}
\def\@onedot{\ifx\@let@token.\else.\null\fi\xspace}

\def\eg{\emph{e.g}\onedot} 
\def\ie{\emph{i.e}\onedot} 
 \def\wrt{w.r.t\onedot}
\def\vs{\emph{vs}\onedot}
\usepackage{newfloat}
\usepackage{listings}
\DeclareCaptionStyle{ruled}{labelfont=normalfont,labelsep=colon,strut=off} 
\floatstyle{ruled}
\newfloat{listing}{tb}{lst}{}
\floatname{listing}{Listing}
\title{Sketch and Refine: Towards Fast and Accurate Lane Detection}
\author{
    Chao Chen,
    Jie Liu\thanks{Corresponding author. liujie@nju.edu.cn},
    Chang Zhou,
    Jie Tang,
    Gangshan Wu
}
\affiliations{
    State Key Laboratory for Novel Software Technology, Nanjing University, China\\
    \{chenchao, zhouchang\}@smail.nju.edu.cn, \{liujie,tangjie,gswu\}@nju.edu.cn
}

\usepackage{bibentry}

\begin{document}

\maketitle

\begin{abstract}
Lane detection is to determine the precise location and shape of lanes on the road. Despite efforts made by current methods, it remains a challenging task due to the complexity of real-world scenarios. Existing approaches, whether proposal-based or keypoint-based, suffer from depicting lanes effectively and efficiently. Proposal-based methods detect lanes by distinguishing and regressing a collection of proposals in a streamlined top-down way, yet lack sufficient flexibility in lane representation. Keypoint-based methods, on the other hand, construct lanes flexibly from local descriptors, which typically entail complicated post-processing. In this paper, we present a “Sketch-and-Refine” paradigm that utilizes the merits of both keypoint-based and proposal-based methods. The motivation is that local directions of lanes are semantically simple and clear. At the “Sketch” stage, local directions of keypoints can be easily estimated by fast convolutional layers. Then we can build a set of lane proposals accordingly with moderate accuracy. At the “Refine” stage, we further optimize these proposals via a novel Lane Segment Association Module (LSAM), which allows adaptive lane segment adjustment. Last but not least, we propose multi-level feature integration to enrich lane feature representations more efficiently. Based on the proposed “Sketch-and-Refine” paradigm, we propose a fast yet effective lane detector dubbed “SRLane”. Experiments show that our SRLane can run at a fast speed (i.e., 278 FPS) while yielding an F1 score of 78.9\%. The source code is available
at: https://github.com/passerer/SRLane.
\end{abstract}

\section{Introduction}

\def\imh{0.9in}
\begin{figure}[ht]
\centering
    \subfigure[]{
\label{figure:illustration:kpt}
 \includegraphics[width = 0.45\linewidth]{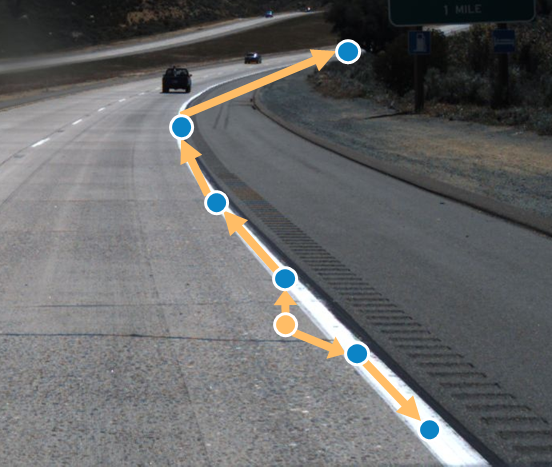}}
    \subfigure[]{
    \label{figure:illustration:anchor}
    \includegraphics[width = 0.45\linewidth]{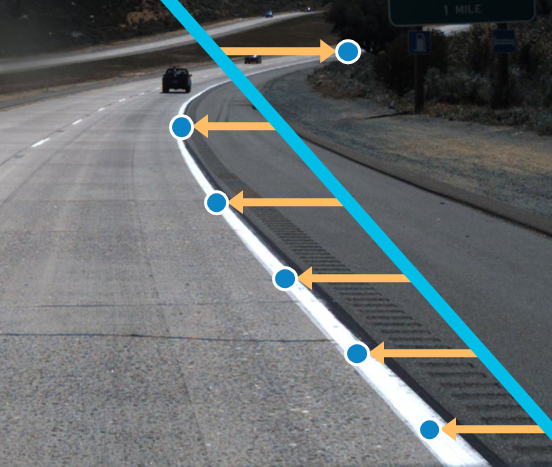}}
\\
     \subfigure[]{
       \label{figure:illustration:ours1}
        \includegraphics[width = 0.45\linewidth]{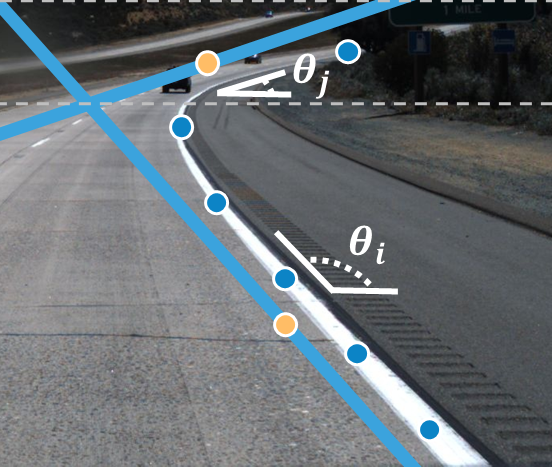}}
    \subfigure[]{
    \label{figure:illustration:ours2}
    \includegraphics[width = 0.45\linewidth]{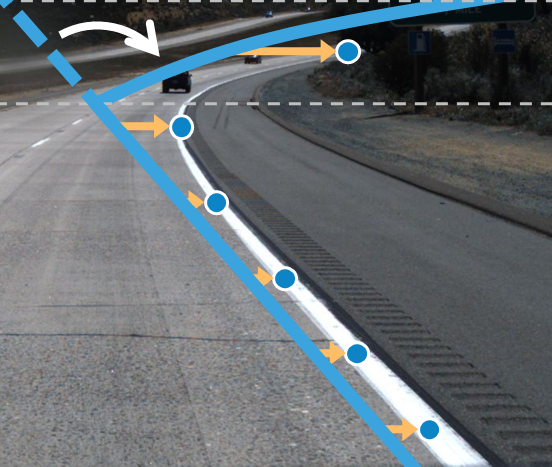}}

\caption{Illustrations of lane detection methods. (a) An example of keypoint-based methods, where local keypoints are grouped based on their offsets (orange colored arrow) from each other to reconstruct the whole lane. (b) An example of anchor-based methods which use pre-defined line anchors (cyan colored line) to match and predict lanes. (c) The proposed paradigm first sketches local directions for a set of keypoints. Then it extends the estimated local directions to build lane proposals.\protect\label{figure:illustration:c} (d) Refinement of lane segments via the proposed Segment Association Module (\protect\ie, the dashed line segment is replaced to better fit the ground truth).} 
\label{figure:illustration}
\end{figure}

Lane detection is a core part of an autonomous driving system that aims to depict the exact shape of each lane on the road. It has advanced thanks to the prosperity of convolutional neural networks (CNN) \cite{scnn,linecnn}. 
Prevailing solutions for lane detection can be broadly divided into two streams: keypoint-based methods 
\cite{folo, ganet}
and proposal-based methods \cite{beziernet,  laneatt}. Keypoint-based methods reduce lane detection into two steps: discrete keypoints detection and association. That is, keypoints on lanes are located first and then heuristically integrated into continuous lane instances via local descriptors. Local descriptors are compact representations of local image characteristics, including semantic ones like deep feature embedding \cite{pinet}, and geometric ones like spatial offsets to adjacent keypoints \cite{folo} (See Fig.\;\ref{figure:illustration:kpt}). Despite their flexibility to describe a variety of lane shapes, the integration procedure complicates the pipeline and causes additive errors. Besides, these methods face difficulty predicting more lane-level attributes, such as lane type. 

Alternatively, proposal-based methods rely on proposals to aggregate relevant features and model global geometry directly. Proposals here refer to a group of class-agnostic candidates likely to be lanes. Taking proposals as input, there is usually a classifier to distinguish between positive and negative samples and a locator to predict shapes. Benefiting from the end-to-end pipeline, these methods are simpler to design and implement. However, the demand for high-quality proposals leaves them in a dilemma between efficiency and adaptability. Typically, sparse proposals are computationally efficient but have limited fitting ability, while dense proposals offer better fitting ability but can be more computationally expensive \cite{ufldv2}.
Some anchor-based methods \cite{linecnn, laneatt} introduce anchors as references to build proposals (See Fig.\;\ref{figure:illustration:anchor}). To guarantee the detection qualities, they preset hundreds of thousands of anchors to cover underlying lanes, which results in high redundancy and complexity and hinders their applications on resource-constrained hardware. For a trade-off between performance and inference speed, some methods \cite{ufld} arrange varying amounts of anchors to different datasets, which involves more laborious processes, while the inherent problem remains untouched.

In this paper, we seek to develop an efficient and effective lane detector by incorporating the characteristics of both paradigms.
As demonstrated in \cite{folo}, local lane markers are easier to predict than global lanes because of their limited geometric variations and spatial coverage. This raises a question: \textit{can we leverage \textbf{cheap} local descriptors to build complex global lane proposals?} We empirically observe that lanes in the front view exhibit certain geometric consistency, \eg, limited variation in lane slope within a local region. 
It implies that keypoints of a lane are likely to lie on or near the extension line of the lane segment. 
Therefore, we are motivated to propose SRLane, where lane shapes are quickly \textbf{S}ketched  (\ie, roughly depicted) using local geometry descriptors and then \textbf{R}efined for higher accuracy. Specifically, our paradigm first predicts a local direction map, wherein the value of each point indicates the approximate direction of the closet lane segment. Then we build the corresponding proposal of each point along its direction
(See Fig.\;\ref{figure:illustration:ours1}). As a result, we can locate lanes fast and fairly accurately. For example, the sketched lanes can reach a detection accuracy of 93\% on Tusimple dataset.

To fit lanes with large curvature variations, a Lane Segment Association Module (LSAM)
is developed to adjust non-fitting segments of lane proposals (See Fig.\;\ref{figure:illustration:ours2}). It is implemented by emphasizing foreground segment features based on the semantic relationships between segments. Moreover, we propose multi-level feature integration tailored for lane detection. 
It directly samples multiple features from different levels of features, which endows a holistic view of lanes.

Such ``Sketch-and-Refine'' paradigm enables us to detect lanes efficiently and effectively.
Extensive experiments on benchmarks demonstrate that our SRLane could obtain remarkable performance with breakneck speed. For example, SRLane can achieve 278 FPS with an F1 score of 78.9\% on CULane test set. We hope SRLane can serve as a new baseline for proposal-based methods and advance the development of real-time lane detection.

\begin{figure*}[tp] 
\begin{center}
   \includegraphics[width=0.98\linewidth]{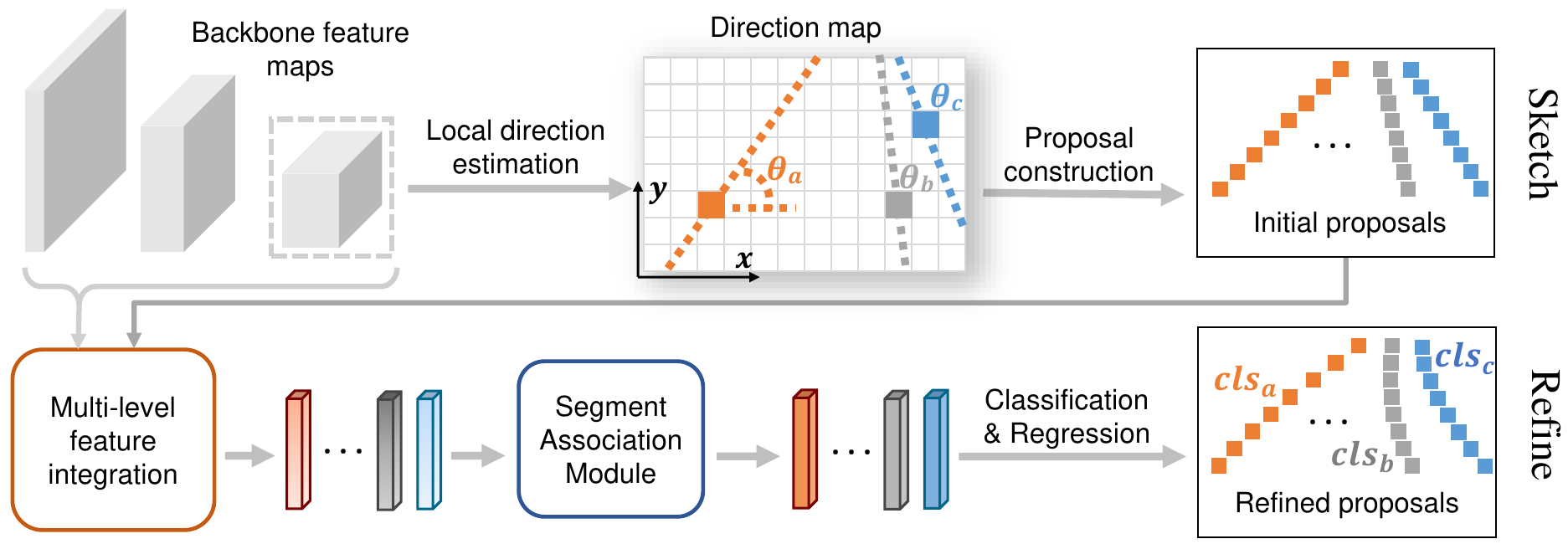}
\end{center}
   \caption{
   Overall pipeline of SRLane, which can be decomposed into two stages: lane sketch and refinement. In the sketch stage, the last feature map from backbone is encoded to create a local direction map,  where a set of lane proposals are initialized. In the refinement stage, features of proposals are adaptively sampled from multi-scale feature maps and then enhanced by the Lane Segment Association Module. After that, they are fed into the classification and regression branch to produce final results.
   }
\label{figure:architecture}
\end{figure*}

\section{Related Work}
Initially, lane detection research mainly focused on detecting hand-crafted low-level features and fitting a spline to localize lanes \cite{ransac, color-based, hmaxima}. Compared to deep learning methods, such methods perform worse in complex scenes. 
Prevailing attempts of deep learning in lane detection can be classified
into two categories: keypoint-based methods and proposal-based methods. Following that,  we provide a succinct description of the representative work for both of these categories.

\textbf{Keypoint-based lane detection.\;}\label{section:keypoint_based}
The core difference of the keypoint-based method is how the keypoint-wise grouping is carried out. PINet \cite{pinet} uses a confidence branch, offset branch, and the embedding branch to cluster keypoints on the lane. FOLOLane \cite{folo} deploys a confidence head and regression head. The regression head estimates the horizontal offsets of each pixel to three neighboring keypoints with fixed vertical intervals. The lane instance takes form by correlating adjacent keypoints. GANet \cite{ganet} predicts all possible keypoints as well, but finds the corresponding lane by adding the coordinate with the offset to the lane line start points. Drawing on the idea of Relay Chain, RCLane \cite{rclane} creates a transfer map and recovers lane instances sequentially in a chain mode.  

Though making use of local descriptors like keypoint-based methods, our proposed SRLane is rooted in the proposal-based method,  which gets rid of complicated post-clustering procedure.

\textbf{Proposal-based lane detection.\;}
Proposal-based methods can be further categorized into two subgroups, \eg, anchor-free methods and anchor-based methods. Among the anchor-free methods, PolyLaneNet \cite{polylane} outputs a small number of proposals from simple fully-connected layers, each of which embraces polynomial coefficients and the confidence score of a lane. BézierLane \cite{beziernet} uses a sparse set of proposals derived from simple column-wise Pooling to predict Bézier curves. Due to simple architecture design and few proposals, PolyLaneNet and BézierLane both run fast. Nevertheless, their performance lags far behind the latest anchor-based methods, which can be attributed in part to the indistinguishable proposal features produced by rough sampling schemes.

As for anchor-based methods, Line-CNN \cite{linecnn} is the pioneering work that puts forward to classify and regress lanes via line anchors. To avoid redundant anchors, SGNet \cite{sgnet} restricts the anchor generation domain to pixels around the vanishing point. To fully utilize line anchors, LaneATT \cite{laneatt} deploys anchor-based pooling for local feature sampling and anchor-based attention for global information fusion. Subsequently, CLRNet \cite{clrnet} refines lane priors hierarchically starting from line anchors, which further boosts performance.
Instead of directly regressing coordinates of lanes, another kind of methods~\cite{condlane, ufld} get lane locations through row-wise or column-wise ordinal classification. However, the performance of above methods is largely affected by the anchor setting. To acquire satisfied detection qualities, they usually drop plenty of anchors, and only a small portion of them become proposals
of interest that contribute to the detection results. In contrast, our method de-emphasizes the number of proposals and achieves higher efficiency.

\section{Methodology}\label{section:method}

\subsection{Overall Pipeline}

As Fig.\;\ref{figure:architecture} shows, the proposed SRLane breaks down the lane detection process into two stages: lane sketch for fast localization of potential lanes and lane refinement for more accurate results. At the sketch stage, SRLane creates a local direction map to initialize lane proposals, which serves as the input to the next stage. At the refinement stage, features of the proposals are dynamically sampled across multiple levels and processed by the Lane Segment Association Module (LSAM) to strengthen segment-level interaction. The refined features are then fed into the classification and regression branches to generate final results. 

All of the parameters in SRLane are trained in an end-to-end supervised manner.
By treating the refinement stage as an RoI head in two-stage detectors \cite{fastercnn}, SRLane can be easily incorporated into other detection frameworks for multi-task detection. In the subsequent sections, we go into the details of these two stages.

\subsection{Lane Sketch}
To provide good proposal initialization in the first stage, we figure out three possible schemes for lane sketch: 
\begin{enumerate*}[label=(\alph*)] 
\item Pre-defining a collection of line anchors like \cite{laneatt}, which is inflexible to fit lines with different curvatures. More importantly, the initial status of anchors is irrelevant to the input image, which deviates from the purpose of the lane sketch.
\item Plugging an intermediate lane proposal network to generate proposals, whose structure can refer to \cite{condlane}. However, capturing lanes' diverse shapes and distribution requires considerable network complexity. Also, the generation process is uninterpretable, which is a barrier to further analysis and improvement.
\item  Using local descriptors, which are preferable since local patterns are relatively simple and easy to predict.
\end{enumerate*}
Constructing lanes from local descriptors has long been researched in keypoint-based methods. Unlike them, we draw on the local continuity of lanes to eliminate the complicated post-clustering procedure and simplify the overall detection pipeline. There are several choices of local descriptors, such as offsets to three neighboring keypoints described in \cite{folo}. For simplicity, we use the coordinates and local direction of the point together to build lane proposals. 

\noindent
\textbf{Local direction estimation.\;}\label{section:local-direction-estimation}
Given an image $\mathbf{I}\in\mathbb{R}^{H\times W\times 3}$, the feature map can be derived from the backbone as $\mathbf{F}^s\in\mathbb{R}^{(H/s)\times (W/s)\times d_s}$, where $s$ is the down-sampling stride and $d_s$ is the channel dimension. We further encode $\mathbf{F}$ to obtain a direction map $\mathbf{\Theta}\in\mathbb{R}^{(H/s)\times (W/s)\times 1}$,  wherein the value of each point represents the local angle of the closest lane, which ranges between $\left[0^{\circ}, 180^{\circ}\right)$. We argue that predicting the overall angle of a lane is inadvisable as it necessitates the global receptive field, and the definition of overall angle is ambiguous if the curvature varies along the lane. Instead, the local direction is semantically simple and clear, allowing it to be estimated even by a single convolutional layer. 

During the training phase, we apply direction supervision on feature maps of all scales,  which is expected to enrich spatial details of the context features. Each annotated lane line is split into $K$ segments, with each segment yielding ground-truth value for its neighboring points. The value of $K$ is given by $K=H/s$ to accommodate different receptive fields.
During inference, only the results of the feature map with the lowest resolution are computed and kept to initialize lane proposals, hence the slight computational cost. 

\noindent
\textbf{Global lane construction.\;}
 Our goal is to build global lane proposals using estimated local geometry descriptors. For point $p$ in the $y$-th row and the $x$-th column of the direction map $\mathbf{\Theta}$, the respective proposal can be formulated as a straight line passing through $p$ with an angle of $\mathbf{\Theta}_{xy}$, as shown in the upper middle part of Fig.\;\ref{figure:architecture}. Following \cite{laneatt},  a lane is represented by a series of lane points. Given a fixed $y_i$, the corresponding $x_i$ of the line point can be inferred by
\begin{equation}
    x_i = \frac{y_i - y}{\text{tan}(\mathbf{\Theta}_{xy})} + x.
\end{equation}

In this way, we can get $(H\times W)/ (s\times s)$ lane proposals. Unlike what is commonly applied in keypoint-based methods \cite{folo, rclane}, we do not estimate a binary classification map to filter out foreground samples
, since the current point's classification result is inadequate for deciding whether its proposal belongs to the foreground.
Also, it is worth noting that the number of proposals can be scaled via the interpolation of $\mathbf{\Theta}$.
Despite the size of $\mathbf{\Theta}$, the derived proposals are always distributed across the entire image plane, each indicating the direction of a potential lane nearby. Therefore, a small number of proposals (\eg, 40)  is sufficient to cover all potential lanes, rather than hundreds of thousands of ones in other works \cite{linecnn, sgnet, laneatt, clrnet}.

\subsection{Lane Refinement}

To accommodate complex shapes of lanes, we develop 
 a refinement head that can adjust lane proposals adaptively. It takes the coarse location of the lane proposal as input and integrates multi-level lane features dynamically. The gathered lane features are then projected into 1D feature vectors. The feature vector is decomposed into groups, each attending to different lane segments. Afterward, we impose Lane Segment Association Module (LSAM) to enhance lane features and feed the updated features into classification and regression branches to generate predictions.

\noindent
\textbf{Multi-level feature integration.\;} 
 Feature integration at multiple feature levels is critical in detection task \cite{yolov4}. Compared to objects represented by rectangular boxes, lane lines are thin and extended with a large span, thus requiring more comprehensive multi-level features. However, how to leverage different feature levels efficiently remains a challenge. In previous research, Feature Pyramid Network (FPN) \cite{fpn} and its variants \cite{pan} are developed to mix low-level and high-level features. Despite their effectiveness, these approaches bring increased complexity.

\begin{figure}[tp] 
\begin{center}
   \includegraphics[width=1.0\linewidth]{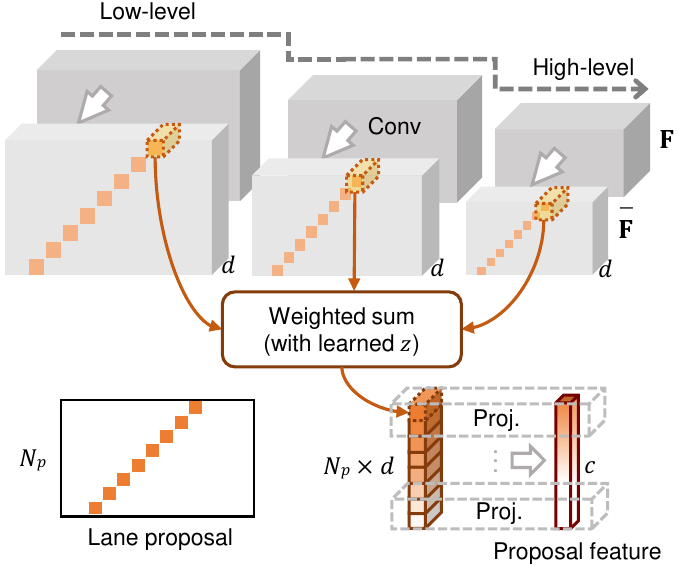}
\end{center}
\caption{Illustration of multi-level feature integration. Before sampling, all feature maps are first transformed by a single convolutional layer to be of the same channel dimension $d$. $N_p$ denotes the number of sample points. The sampled features are finally projected to a 1D feature vector with channel $c$.}
\label{figure:adaptive_sampling}
\end{figure}

To this end, we forego the heavy FPN in favor of sampling multi-level features directly and adaptively. As shown in Fig.\;\ref{figure:adaptive_sampling}, we sample features of $N_p$ points in the lane and expand the sampling coordinate of $i$-th point from $\left(x_i, y_i\right)$ to $\left(x_i, y_i, z_i\right)$, where the scalar $z_i$ represents the feature scale of greater significance. The sampling process is completed softly, with the sampled features being a Gaussian weighted sum over scales:
\begin{align}
    \mathbf{x}  = \text{Proj}\left(\left\{\sum_s \overline{\mathbf{F}}^s_{x_i y_i} \cdot \frac{\text{Exp}(-|2^{z_i}-s|)}{\sum_{s^{'}} \text{Exp}(-|2^{z_i}-s^{'}|)}\right\}_{i=0}^{N_{p}-1} \right).
\end{align}
Here $\overline{\mathbf{F}}^s$ is the feature map with stride $s$. Zero values will be padded if $(x_i, y_i)$ falls outside the feature map. Large $z_i$ implies that high-level features are desired while low-level features are suppressed, and vice versa. $\text{Proj}(\cdot)$ is the projection function, and $\mathbf{x}\in \mathbb{R}^c$ is the projected feature of the lane proposal. We divide sample points into $G$ groups along the vertical axis and map features of the same group to the corresponding channel group of $\mathbf{x}$.

Unlike traditional sampling strategies, adaptive sampling over scales can reduce redundant features and exploit more valuable information. As a result,  the output lane features where multi-level clues co-exist serve as a solid foundation for the following segment association module. It is also desirable for classification and localization, as it requires high-level semantic information and low-level location details.

In practice, we make the vector $\textbf{z}=\left[z_0, \cdots, z_{N_p-1} \right]$ learnable, which is known as \textit{embedding}. We discover that the convergence of $\textbf{z}$ is insensitive to the initial state, which indicates the successful modeling of underlying sampling patterns.

\begin{figure}[tp] 
\begin{center}
   \includegraphics[width=0.9\linewidth]{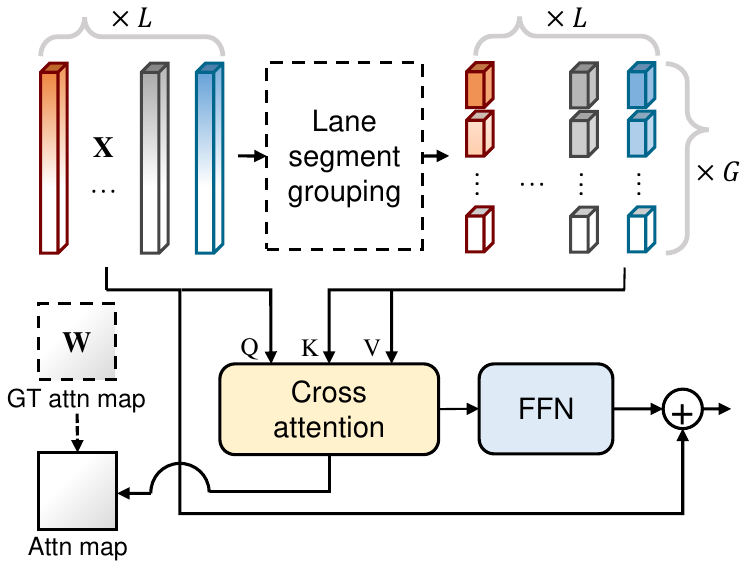}
\end{center}
\caption{Illustration of Lane Segment Association Module. $G$ denotes the number of groups and $L$ is the number of lane proposals. \texttt{FFN} is the Feed Forward Network \cite{detr}. For simplicity, the batch dimension is ignored.
}
\label{figure:msa}
\end{figure}

\begin{table*}[tp]
\begin{center}
\resizebox{\linewidth}{!}{
\begin{tabular}
{l p{0.6cm}<{\centering}c|cl cccccccccc}
\hline
Method & Year & Backbone  &  \multicolumn{2}{c}{Latency} & Normal & Crowd & Dazzle & Shadow & No line & Arrow & Cross & Night  & Curve &  Total  \\
\hline
\multicolumn{14}{l}{{keypoint-based}}\\ \hline
LaneAF & 2021 & DLA34 &  & \; -   & 91.8 & 75.6 & 71.8 & 79.1 & 51.4 & 86.9  & 1360 & 73.0 & 72.0 & 77.4 \\
FOLOLane & 2021 & ERFNet & & 40$^\top$    & 92.7 & 77.8 & \textbf{75.2} & 79.3 & 52.1 & 89.0 & 1569 & \textbf{74.5} & 69.4 & 78.8 \\
RCLane & 2022 & SegB0 & & 22$^\top$  & \textbf{93.4} & \textbf{77.9} & 73.3 & \textbf{80.3} & \textbf{53.8} & 89.0 & 1298 & 74.3  & 75.7 & \textbf{79.5}  \\
GANet & 2022 & Res18 & & \textbf{7.5} & 93.2 & 77.2& 71.2 & 77.9 & 53.6 & \textbf{89.6} & \textbf{1240} & 72.8  & \textbf{75.9} & 78.8  \\
\hline
\multicolumn{14}{l}{{proposal-based}}\\ \hline
SGNet & 2021 & Res18 & & 8.5$^\top$   & 91.4 & 74.0 & 66.9 & 72.2 & 50.2 & 87.1  & 1164 & 70.7 & 67.0 & 76.1  \\
CLRNet & 2022 & Res18 & & 8.0  & 93.3 & \textbf{78.3} &\textbf{73.7} & 79.7 & \textbf{53.1} & \textbf{90.3}  & 1321 & \textbf{75.1}  & 71.6 & \textbf{79.6} \\
UFLDv2 & 2022 & Res18 & & 6.5   & 91.8 & 73.3 & 65.3 & 75.1 & 47.6 & 87.9 & 2075 & 70.7 & 68.5 & 75.0 \\ 
CondLane & 2021 & Res18 & & 6.1   & 92.9 & 75.8 & 70.7 & \textbf{80.0} & 52.4 & 89.4  & 1364 & 73.2 & 72.4 & 78.1 \\
\hdashline 
LaneATT & 2021 & Res18 & & 3.7   & 91.2 & 72.7 & 65.8 & 68.0 & 49.1 & 87.8  & 1020 & 68.6 & 63.8 & 75.1 \\
BézierLane & 2022 & Res18 & & \textbf{3.6}   & 90.2 & 71.6 & 62.5 & 70.9 & 45.3 & 84.1  & \textbf{996} & 68.7 & 59.0 & 73.8 \\ 
\textbf{Ours} & & Res18 & & \textbf{3.6}  &  \textbf{93.5} & 77.8  &  71.6 &  78.8 & 52.1  & 90.2  & 1365  & 74.7  & \textbf{74.7} & 78.9 \\
\hline
\end{tabular}
}
\end{center}
\caption{Comparison of F1-measure and inference latency on CULane \protect\texttt{test} set with state-of-the-art methods. The unit of latency is ``ms". For the ``Cross" category, the number of false positives is recorded. Superscript ``$\top$'' indicates the latency is not re-measured under the same conditions since the source codes are unavailable, and we use the data reported in the paper. The best results are marked bold.\protect\label{table:culanesota}}
\end{table*}

\begin{table}[tp]
\begin{center}
\resizebox{1.0\linewidth}{!}{
\begin{tabular}{l|cl cccc}
\hline
Method  & \multicolumn{2}{c}{\makecell{Latency$\downarrow$ \\ (ms)}}  & \makecell{Acc$\uparrow$\\(\%)} & \makecell{F1$\uparrow$\\(\%)} & \makecell{FP$\downarrow$\\(\%)} & \makecell{FN$\downarrow$\\(\%)} \\
\hline
\multicolumn{6}{l}{{keypoint-based}}\\ \hline
LaneAF-DLA34 & & \; -  & 95.62 & 96.49 & 2.80 & 4.18 \\
FOLOLane-ERFNet & & $40^\top$  & \textbf{96.92} & 96.59 & 4.47 & \textbf{2.28} \\
RCLane-SegB0 & & $22^\top$  & 96.58  &  97.64 & 2.21 & 2.57  \\
GANet-Res18 & & \textbf{7.5}  &  95.95 & \textbf{97.71} &  \textbf{1.97} & 2.62 \\
\hline
\multicolumn{6}{l}{{proposal-based}}\\ \hline
CLRNet-Res18 & & 8.0 &  96.84 &  \textbf{97.89} & 2.28  &  1.92  \\
CondLane-Res18 & & 6.1 &  95.48 & 97.01  & \textbf{2.18}  & 3.80 \\
UFLDv2-Res18 & & 4.0  & 95.65 & 96.16 & 3.06 & 4.61 \\
LaneATT-Res18 & & 3.7 & 95.57  &  96.71 & 3.56  & 3.01   \\
\textbf{Ours}-Res18 & & \textbf{3.6} & \textbf{96.85} & 97.66 & 2.80 & \textbf{1.85} \\
\hline
\end{tabular}
}
\end{center}
\caption{Comparison with state-of-the-art methods on Tusimple test set. \protect\texttt{Acc} is the abbreviation of accuracy. $\downarrow$ in the head row indicates that the lower the metric is, the better the performance is, and vice versa.\protect\label{table:tusimplesota}}
\end{table}

\noindent
\textbf{Lane segment association.\;} \label{section:msa}
The sketch stage provides a set of lane proposals that roughly depict the actual lane, while certain positions may need to be adjusted. For this purpose, we propose a novel Lane Segment Association Module (LSAM) for segment-level refinement. Given features of $L$ lane proposals $\mathbf{X}=\left[\mathbf{x}_0, \cdots, \mathbf{x}_{L-1}\right]\in \mathbb{R}^{L\times c}$, each $\mathbf{x}_i \in \mathbb{R}^{c}$ represents the entire lane feature. Meanwhile, it can also be regarded as an ordered collection of $G$ lane segment features: $\mathbf{x}_i = [\mathbf{x}_i^0, \cdots, \mathbf{x}_i^{G-1}]$, where $\mathbf{x}_i^g \in \mathbb{R}^{c/G}$. As shown in Fig.\;\ref{figure:msa}, each proposal feature $\mathbf{x}_i$ acts as a \textit{query} to gather information from all $L\times G$ segment features adaptively by cross-attention mechanism. It is expected that the foreground-like proposals could collect hints from foreground-like segments, obtaining a high tolerance for lane variations. The proposed LSAM is de-facto computationally friendly thanks to the small number of proposals.

Although LSAM reinforces the communication between semantically similar (\eg, both foreground-like) instances, it may fall into a suboptimal process without explicit supervision. To mitigate this issue, we use bipartite matched ground truth (GT) to supervise the attention weights in LSAM. Given any pair of proposals denoted as $(l_i, l_j)$, their target attention weight of $g$-th group $\mathbf{W}_{(i,j)}^g$ is defined as:
\begin{equation}
    \mathbf{W}_{(i,j)}^g = 
    \left\{
\begin{aligned}
1 \quad & if\  j=\mathop{\arg\min}\limits_{k}\text{d}\left(l_k^g, \hat{l}_i^g\right) \\ 
0 \quad & \text{otherwise}
\end{aligned}
\right. 
,
\end{equation}
where $\hat{l}_i$ is the matched GT of ${l}_i$, and $\text{d}\left(\cdot, \cdot\right)$ measures the geometry distance between two lane segments. Supervised by the cross entropy loss of attention weights, each proposal is forced to attend to its target segment and learn better feature representation.

\noindent
\textbf{Classification and regression.\;}
Taking proposal features as input, the classification branch predicts the probability of the proposal being a foreground, while the regression branch yields a more precise location.
The location of a lane is described by a series of x-coordinates of lane points with equal vertical intervals, which is denoted as $\left\{ x_i \right\}_0^{N-1}$, where $N$ represents the total number of points. The y-coordinate corresponds to $x_i$ is given by $y_i = i\cdot \frac{H}{N-1}$, where $H$ is the height of image. Besides, the regression branch also predicts the maximum and minimum y-coordinate values to determine the endpoints of the lane.

\section{Experiment}

\subsection{Datasets and Evaluation Metrics}\label{section:dataset}

Experiments are conducted on three popular lane datasets in the literature: Tusimple \cite{tusimple}, CULane \cite{scnn}, and Curvelanes \cite{curvelane}.
Tusimple contains 3,626 images for training and 2,782 images for testing, all collected in highway scenes.
CULane is one of the largest lane detection datasets, which comprises 88,880 frames, 9,675 frames, 34,680 frames for training, validation, and testing, respectively. 
CurveLanes is a recently released benchmark with dense curve lanes, and we use it's subset for hyperparameter tuning and ablation experiments.

In line with the official metric used in \cite{tusimple} and \cite{scnn}, we use accuracy for Tusimple and F1 for CULane as the main evaluation metrics. F1 is a holistic metric that combines true positive (TP), false positive (FP), and false negative (FN). Besides accuracy, we also report F1, FP, and FN ratios on Tusimple. 

\subsection{Implementation Details}\label{section:implementation}

\noindent
\textbf{Architecture.\;}
We adopt a standard ResNet18 \cite{resnet} as the pre-trained backbone and use multi-scale feature maps from the last three stages. 
Currently, the local direction map is resized to $4\times 10$ regardless of the input resolution, which means the number of proposals $L$ is fixed as 40. 
The groups of lane segments is set to 6 by default. In the current implementation, all operators in the model are based on PyTorch \cite{pytorch}.

\noindent
\textbf{Loss.\;}
Training loss encompasses $l_1$ loss for direction estimation, cross entropy loss for attention weights, focal loss \cite{focalloss} for lane classification, and iou loss  \cite{clrnet} for lane regression. The overall loss is given by:
\begin{equation}
    \mathcal{L} = {w}_{cls} \mathcal{L}_{cls} + {w}_{reg} \mathcal{L}_{reg} + w_{dir}   \mathcal{L}_{dir} + w_{attn}\mathcal{L}_{attn},
\end{equation}
where the loss weights are set as $w_{cls}=2.0$, $w_{reg}=1.0$, $w_{dir}=0.05$, and $w_{attn}=0.05$.
The direction loss is only performed on points adjacent to ground truth within a certain range. The lane regression loss is performed on the positive samples which have the highest matching degree with ground truth.

\noindent
\textbf{Training.\;}
We use AdamW \cite{adamw} as the optimizer in conjunction with a cosine learning rate scheduler. The initial learning rate is set to $10^{-3}$ with 800 iterations of linear warm-up. The batch size is set to 40 and images are resized to $800\times 320$. Data augmentation for training includes random flipping, affine transformation, color jittering and JPEG compression. All models can be trained on a single GPU with 12GB memory.

\begin{figure}[tp] 
\begin{center}
   \includegraphics[width=1.0\linewidth, trim=0 0 0 0, clip]{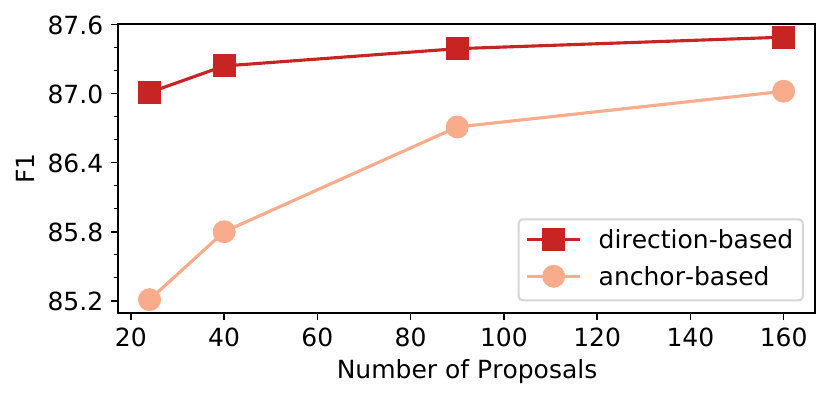}
\end{center}
\caption{Performance comparison  with different number of proposals.}
\label{figure:number_ablation}
\end{figure}

\begin{figure*}[tp] 
\begin{center}
   \includegraphics[width=0.98\linewidth]{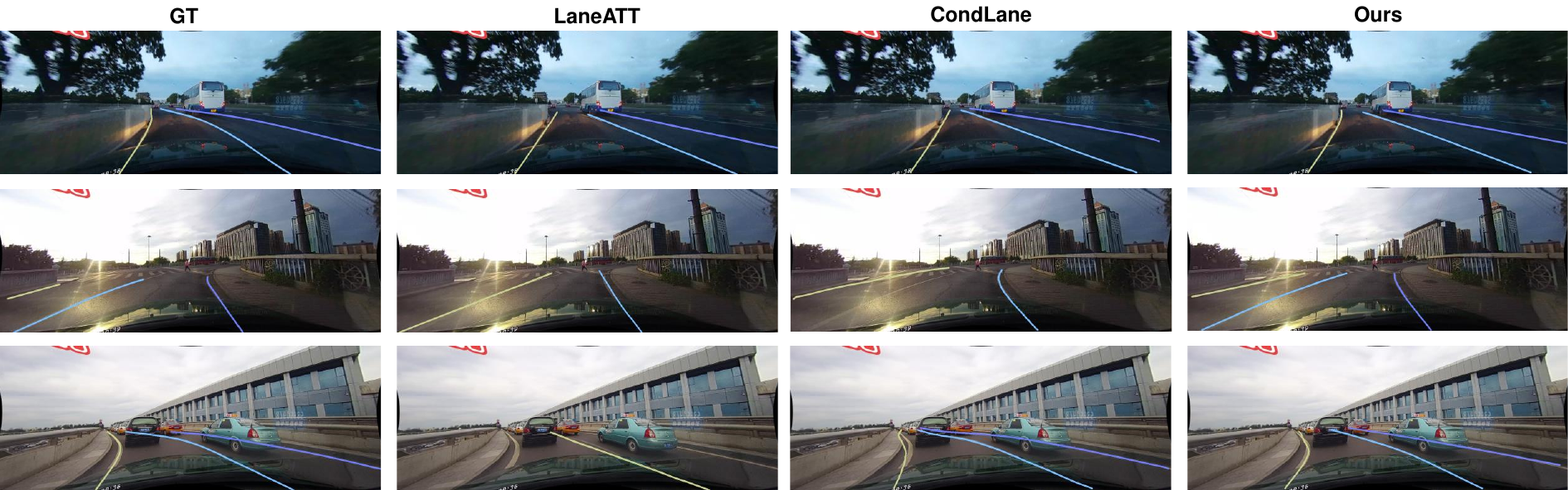}
\end{center}
\caption{Qualitative comparison between our SRLane and other lane detection methods (Best viewed in colors).
}
\label{figure:detection_vis}
\end{figure*}

\subsection{Main Results}

We primarily focus on the comparison with cutting-edge methods \cite{laneaf, beziernet, condlane, ufldv2, folo, laneatt, ganet, rclane,clrnet}, which are summarized into two categories: keypoint-based methods and proposal-based methods. For a fair comparison, the inference latency is measured on our machine with an AMD EPYC 7232P CPU and an NVIDIA Titan Xp GPU.

\noindent
\textbf{CULane.\;}
 The performance of SRLane on the CULane dataset is compared against other methods in Tab.\;\ref{table:culanesota}. We can see that SRLane could run extremely fast and get superior results in total compared to those of similar speed, \eg, 3.8\% F1 gain \vs LaneATT and 5.1\% F1 gain \vs BézierLane.
In curve scenarios, the benchmarks are dominated by keypoint-based detectors, \eg, GANet achieves 75.92\% F1. Notably, SRLane obtains a competitive 74.7\% F1 on the curve set, narrowing the gap between state-of-the-art keypoint-based detectors and surpassing the previous highest curve score of proposal-based detectors by 2.3\%. It demonstrates the promising ability of SRLane to fit curve lanes. 

\noindent
\textbf{Tusimple.\;}
We also run experiments on the Tusimple dataset. As listed in Tab.\;\ref{table:tusimplesota}, our SRLane achieves the fastest speed, which is 2.2 times faster than CLRNet and 11.1 times faster than FOLOLane. Meanwhile, SRLane maintains high accuracy, which demonstrates the superiority of our method.

\noindent
\textbf{Visualization of detection results.\;}
Fig.\;\ref{figure:detection_vis} exemplifies the qualitative results of our method. It can be observed that SRLane achieves a smoother fit to lane curves, especially curve segments near the vanishing point. In addition, SRLane is robust to different lighting conditions, such as dim and dazzling.

\begin{table}[t]
\begin{center}
\centering

\begin{tabular}[t]{c|c|c|c}
\hline
 & CLRNet & UFLDv2 & \textbf{SRLane} \\
\hline
Param. (M) & 0.4 & 85.2 & 0.4\\
MACs (M) & 2450 & 85 & 13 \\ \hline
\end{tabular}
\end{center}
\caption{Model head parameters and MACs. The input image size for MACs is $800 \times 320$.}
\label{table:params_macs}
\end{table}

\noindent
\textbf{Parameters and MACs.\;} 
 Tab.\;\ref{table:params_macs} compares parameters and multiply accumulate operations (MACs) of our SRLane to those of two other prominent models:  CLRNet \cite{clrnet} and UFLDv2 \cite{ufldv2}.  The backbones of all models are  ResNet18, and only the data of head are displayed for clearer comparison. It is shown that SRLane has a significant edge considering both aspects, demonstrating the great practical value in constrained vehicle computing devices.
 
\begin{table}[t]
\begin{center}
\centering
\begin{tabular}[t]{cc|c}
\hline
 Adaptive Sampling & Segment Association &  F1  \\
\hline
 &  & $85.7$  \\
\checkmark  &  & $86.3$  \\
 &  \checkmark  & $86.5$  \\
 \checkmark & \checkmark& $87.2$  \\
\hline

\end{tabular}
\end{center}
\caption{Effects of each component in the proposed SRLane. \protect\label{table:ab_study}}
\label{table:msa}
\end{table}

\subsection{Ablation Study}
\noindent
\textbf{Proposal initialization.\;}
We first verify the necessity of the direction-based proposal initialization. We use the same number of line anchors to initialize lane proposals as the counterpart. The anchor setting is in accord with \cite{clrnet}. As illustrated in Fig.\;\ref{figure:number_ablation}, the proposed direction-based initialization approach achieves excellent performance with 40 proposals, outperforming the corresponding anchor-based one, which needs far more proposals to compensate for the performance gap. 
These findings suggest that the proposed direction-based initialization method can achieve high performance with sparse proposals.

\noindent
\textbf{Multi-level feature integration.\;}
To ablate the effectiveness of our proposed feature sampling strategy, we set the single-level feature sampling method as the baseline. For a fair comparison, we equip the baseline model with Feature Pyramid Network (FPN) to compensate for multi-level information. As shown in Tab.\;\ref{table:ab_study}, using single-level features leads to inferior performance (See row 1 \vs row 2, row 3 \vs row 4), indicating the benefit of sampling multi-level features adaptively. We refer readers to the Appendix for more ablations about it.

\begin{figure}[tp] 
\begin{center}
   \includegraphics[width=0.95\linewidth]{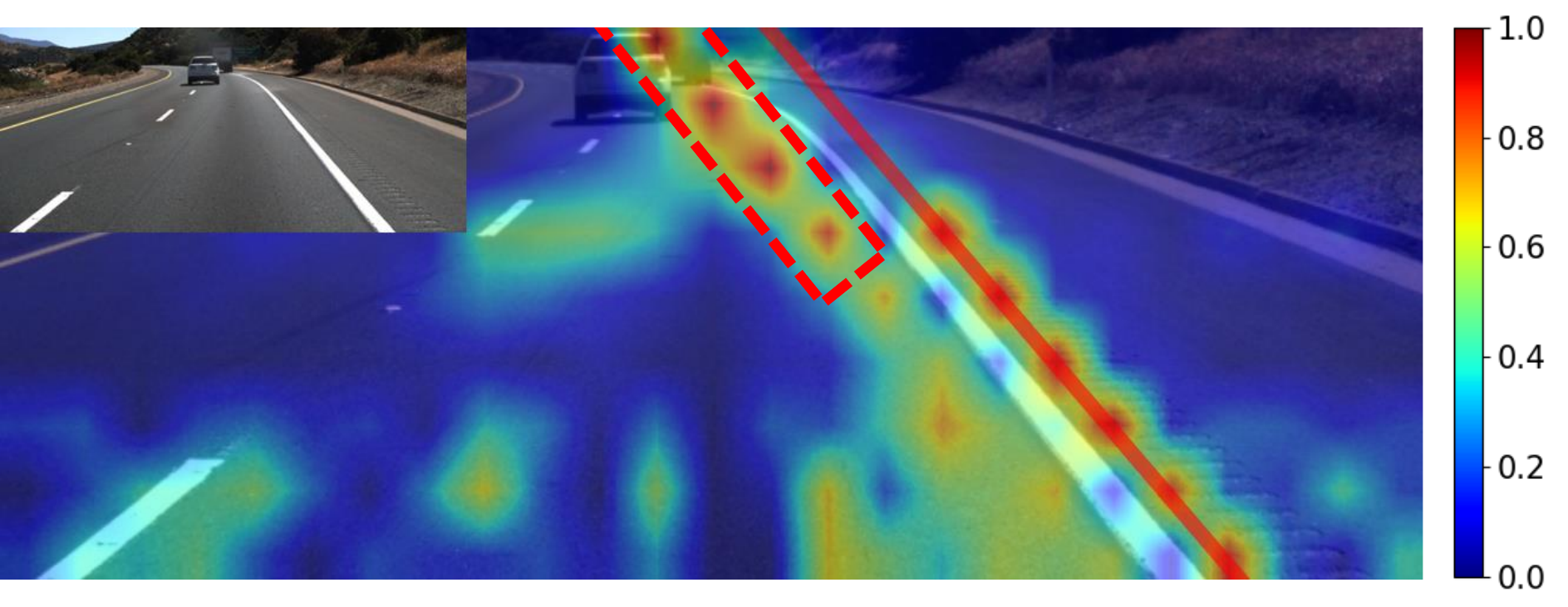}
\end{center}
\caption{Attention weight \wrt the red-colored line proposal. For better visualization, we spread the original segment-level attention weight onto the image plane based on the position of segments. The original image is shown in the upper left.}
\label{figure:msa-weight}
\end{figure}

\noindent
\textbf{Lane segment association module.\;}
We also validate the effectiveness of our proposed  Lane Segment Association Module (LSAM) in Tab.\;\ref{table:ab_study}. It can be seen that LSAM gives an improvement by up to 0.9\% F1 (See row 2 \vs row 4).
To further validate the effectiveness of LSAM, we visualize the attention map in Fig.\;\ref{figure:msa-weight}. LSAM successfully learns to refine segment features by assigning higher attention values to the segment of interest, \eg, the segment closest to the exact foreground, which is enclosed by a red dotted box in the figure. These results demonstrate that LSAM enables a better fit for lanes with significant curvature variations.

\section{Conclusion}

In this paper, we have presented a novel method for the task of 2D lane detection, which allows us to locate lanes accurately and efficiently. It is achieved by roughly sketching the shape
of the lane conditioned on local geometry descriptors and refining it progressively.
Benchmarking on different datasets demonstrates
the outstanding speed, accuracy, and robustness of our SRLane to diverse scenarios. We believe that such a design is transferable to 3D lane detection. However, it is beyond the purpose of this work.

\section{Acknowledgments}
This work is supported by National
Key R\&D Program of China (No. 2022ZD0160900), National Natural Science Foundation of China (No. 62076119,
No. 61921006, No. 62072232), Fundamental Research
Funds for the Central Universities (No. 020214380091,
No. 020214380099), and Collaborative Innovation Center
of Novel Software Technology and Industrialization.

\bibliography{aaai24}

\begin{thebibliography}{52}
\providecommand{\natexlab}[1]{#1}

\bibitem[{Abualsaud et~al.(2021)Abualsaud, Liu, Lu, Situ, Rangesh, and Trivedi}]{laneaf}
Abualsaud, H.; Liu, S.; Lu, D.~B.; Situ, K.; Rangesh, A.; and Trivedi, M.~M. 2021.
\newblock Laneaf: Robust multi-lane detection with affinity fields.
\newblock \emph{IEEE Robotics and Automation Letters}, 6(4): 7477--7484.

\bibitem[{Bochkovskiy, Wang, and Liao(2020)}]{yolov4}
Bochkovskiy, A.; Wang, C.-Y.; and Liao, H.-Y.~M. 2020.
\newblock Yolov4: Optimal speed and accuracy of object detection.
\newblock \emph{arXiv preprint arXiv:2004.10934}.

\bibitem[{Borkar, Hayes, and Smith(2009)}]{ransac}
Borkar, A.; Hayes, M.; and Smith, M.~T. 2009.
\newblock Robust lane detection and tracking with ransac and kalman filter.
\newblock In \emph{2009 16th IEEE International Conference on Image Processing}, 3261--3264.

\bibitem[{Cai and Vasconcelos(2018)}]{casecadercnn}
Cai, Z.; and Vasconcelos, N. 2018.
\newblock Cascade r-cnn: Delving into high quality object detection.
\newblock In \emph{Proceedings of the IEEE Conference on Computer Vision and Pattern Recognition}, 6154--6162.

\bibitem[{Carion et~al.(2020)Carion, Massa, Synnaeve, Usunier, Kirillov, and Zagoruyko}]{detr}
Carion, N.; Massa, F.; Synnaeve, G.; Usunier, N.; Kirillov, A.; and Zagoruyko, S. 2020.
\newblock End-to-end object detection with transformers.
\newblock In \emph{Computer Vision--ECCV 2020: 16th European Conference, Glasgow, UK, August 23--28, 2020, Proceedings, Part I 16}, 213--229.

\bibitem[{Chiu and Lin(2005)}]{color-based}
Chiu, K.-Y.; and Lin, S.-F. 2005.
\newblock Lane detection using color-based segmentation.
\newblock In \emph{IEEE 2005 Intelligent Vehicles Symposium}, 706--711.

\bibitem[{Feng et~al.(2022{\natexlab{a}})Feng, Guo, Tan, Xu, Wang, and Ma}]{beziernet}
Feng, Z.; Guo, S.; Tan, X.; Xu, K.; Wang, M.; and Ma, L. 2022{\natexlab{a}}.
\newblock Rethinking efficient lane detection via curve modeling.
\newblock In \emph{Proceedings of the IEEE/CVF Conference on Computer Vision and Pattern Recognition}, 17062--17070.

\bibitem[{Feng et~al.(2022{\natexlab{b}})Feng, Guo, Tan, Xu, Wang, and Ma}]{beziernet_simp}
Feng, Z.; Guo, S.; Tan, X.; Xu, K.; Wang, M.; and Ma, L. 2022{\natexlab{b}}.
\newblock Rethinking efficient lane detection via curve modeling.
\newblock In \emph{CVPR}.

\bibitem[{Fleuret and Geman(2001)}]{coarsetofinefacedetection}
Fleuret, F.; and Geman, D. 2001.
\newblock Coarse-to-fine face detection.
\newblock \emph{International Journal of computer vision}, 41(1-2): 85.

\bibitem[{Gao et~al.(2022)Gao, Wang, Han, and Guo}]{adamixer}
Gao, Z.; Wang, L.; Han, B.; and Guo, S. 2022.
\newblock Adamixer: A fast-converging query-based object detector.
\newblock In \emph{Proceedings of the IEEE/CVF Conference on Computer Vision and Pattern Recognition}, 5364--5373.

\bibitem[{Ghazali, Xiao, and Ma(2012)}]{hmaxima}
Ghazali, K.; Xiao, R.; and Ma, J. 2012.
\newblock Road lane detection using H-maxima and improved hough transform.
\newblock In \emph{2012 Fourth International Conference on Computational Intelligence, Modelling and Simulation}, 205--208.

\bibitem[{Ghiasi, Lin, and Le(2019)}]{nasfpn}
Ghiasi, G.; Lin, T.-Y.; and Le, Q.~V. 2019.
\newblock Nas-fpn: Learning scalable feature pyramid architecture for object detection.
\newblock In \emph{Proceedings of the IEEE/CVF Conference on Computer Vision and Pattern Recognition}, 7036--7045.

\bibitem[{Gong, Zhao, and Li(2019)}]{ipr}
Gong, J.; Zhao, Z.; and Li, N. 2019.
\newblock Improving Multi-stage Object Detection via Iterative Proposal Refinement.
\newblock In \emph{BMVC}, 223.

\bibitem[{Han et~al.(2022)Han, Deng, Cai, Yang, Xu, Xu, and Liang}]{laneformer}
Han, J.; Deng, X.; Cai, X.; Yang, Z.; Xu, H.; Xu, C.; and Liang, X. 2022.
\newblock Laneformer: Object-aware Row-Column Transformers for Lane Detection.
\newblock In \emph{Proceedings of the AAAI Conference on Artificial Intelligence}, 799--807.

\bibitem[{He et~al.(2017)He, Gkioxari, Doll{\'a}r, and Girshick}]{maskrcnn}
He, K.; Gkioxari, G.; Doll{\'a}r, P.; and Girshick, R. 2017.
\newblock Mask r-cnn.
\newblock In \emph{Proceedings of the IEEE/CVF International Conference on Computer Vision}, 2961--2969.

\bibitem[{He et~al.(2016)He, Zhang, Ren, and Sun}]{resnet}
He, K.; Zhang, X.; Ren, S.; and Sun, J. 2016.
\newblock Deep residual learning for image recognition.
\newblock In \emph{Proceedings of the IEEE Conference on Computer Vision and Pattern Recognition}, 770--778.

\bibitem[{Hou et~al.(2019)Hou, Ma, Liu, and Loy}]{sad}
Hou, Y.; Ma, Z.; Liu, C.; and Loy, C.~C. 2019.
\newblock Learning lightweight lane detection cnns by self attention distillation.
\newblock In \emph{Proceedings of the IEEE/CVF International Conference on Computer Vision}, 1013--1021.

\bibitem[{Hur, Kang, and Seo(2013)}]{randomfileds}
Hur, J.; Kang, S.-N.; and Seo, S.-W. 2013.
\newblock Multi-lane detection in urban driving environments using conditional random fields.
\newblock In \emph{2013 IEEE Intelligent Vehicles Symposium}, 1297--1302.

\bibitem[{Ko et~al.(2021{\natexlab{a}})Ko, Lee, Azam, Munir, Jeon, and Pedrycz}]{kppis}
Ko, Y.; Lee, Y.; Azam, S.; Munir, F.; Jeon, M.; and Pedrycz, W. 2021{\natexlab{a}}.
\newblock Key points estimation and point instance segmentation approach for lane detection.
\newblock \emph{IEEE Transactions on Intelligent Transportation Systems}, 23(7): 8949--8958.

\bibitem[{Ko et~al.(2021{\natexlab{b}})Ko, Lee, Azam, Munir, Jeon, and Pedrycz}]{pinet}
Ko, Y.; Lee, Y.; Azam, S.; Munir, F.; Jeon, M.; and Pedrycz, W. 2021{\natexlab{b}}.
\newblock Key points estimation and point instance segmentation approach for lane detection.
\newblock \emph{IEEE Transactions on Intelligent Transportation Systems}, 23(7): 8949--8958.

\bibitem[{Lee et~al.(2017)Lee, Kim, Shin~Yoon, Shin, Bailo, Kim, Lee, Seok~Hong, Han, and So~Kweon}]{vpgnet}
Lee, S.; Kim, J.; Shin~Yoon, J.; Shin, S.; Bailo, O.; Kim, N.; Lee, T.-H.; Seok~Hong, H.; Han, S.-H.; and So~Kweon, I. 2017.
\newblock Vpgnet: Vanishing point guided network for lane and road marking detection and recognition.
\newblock In \emph{Proceedings of the IEEE/CVF International Conference on Computer Vision}, 1947--1955.

\bibitem[{Li et~al.(2019)Li, Li, Hu, and Yang}]{linecnn}
Li, X.; Li, J.; Hu, X.; and Yang, J. 2019.
\newblock Line-cnn: End-to-end traffic line detection with line proposal unit.
\newblock \emph{IEEE Transactions on Intelligent Transportation Systems}, 21(1): 248--258.

\bibitem[{Lin et~al.(2017{\natexlab{a}})Lin, Doll{\'a}r, Girshick, He, Hariharan, and Belongie}]{fpn}
Lin, T.-Y.; Doll{\'a}r, P.; Girshick, R.; He, K.; Hariharan, B.; and Belongie, S. 2017{\natexlab{a}}.
\newblock Feature pyramid networks for object detection.
\newblock In \emph{Proceedings of the IEEE Conference on Computer Vision and Pattern Recognition}, 2117--2125.

\bibitem[{Lin et~al.(2017{\natexlab{b}})Lin, Goyal, Girshick, He, and Doll{\'a}r}]{focalloss}
Lin, T.-Y.; Goyal, P.; Girshick, R.; He, K.; and Doll{\'a}r, P. 2017{\natexlab{b}}.
\newblock Focal loss for dense object detection.
\newblock In \emph{Proceedings of the IEEE/CVF International Conference on Computer Vision}, 2980--2988.

\bibitem[{Liu et~al.(2021)Liu, Chen, Zhu, and Tan}]{condlane}
Liu, L.; Chen, X.; Zhu, S.; and Tan, P. 2021.
\newblock Condlanenet: a top-to-down lane detection framework based on conditional convolution.
\newblock In \emph{Proceedings of the IEEE/CVF International Conference on Computer Vision}, 3773--3782.

\bibitem[{Liu et~al.(2018)Liu, Qi, Qin, Shi, and Jia}]{pan}
Liu, S.; Qi, L.; Qin, H.; Shi, J.; and Jia, J. 2018.
\newblock Path aggregation network for instance segmentation.
\newblock In \emph{Proceedings of the IEEE Conference on Computer Vision and Pattern Recognition}, 8759--8768.

\bibitem[{Loshchilov and Hutter(2017)}]{adamw}
Loshchilov, I.; and Hutter, F. 2017.
\newblock Decoupled weight decay regularization.
\newblock \emph{arXiv preprint arXiv:1711.05101}.

\bibitem[{Neven et~al.(2018)Neven, De~Brabandere, Georgoulis, Proesmans, and Van~Gool}]{isa}
Neven, D.; De~Brabandere, B.; Georgoulis, S.; Proesmans, M.; and Van~Gool, L. 2018.
\newblock Towards end-to-end lane detection: an instance segmentation approach.
\newblock In \emph{2018 IEEE Intelligent Vehicles Symposium}, 286--291.

\bibitem[{Pan et~al.(2018)Pan, Shi, Luo, Wang, and Tang}]{scnn}
Pan, X.; Shi, J.; Luo, P.; Wang, X.; and Tang, X. 2018.
\newblock Spatial as deep: Spatial cnn for traffic scene understanding.
\newblock In \emph{Proceedings of the AAAI Conference on Artificial Intelligence}.

\bibitem[{Paszke et~al.(2019)Paszke, Gross, Massa, Lerer, Bradbury, Chanan, Killeen, Lin, Gimelshein, Antiga et~al.}]{pytorch}
Paszke, A.; Gross, S.; Massa, F.; Lerer, A.; Bradbury, J.; Chanan, G.; Killeen, T.; Lin, Z.; Gimelshein, N.; Antiga, L.; et~al. 2019.
\newblock Pytorch: An imperative style, high-performance deep learning library.
\newblock \emph{Advances in Neural Information Processing Systems}, 32.

\bibitem[{Qin, Wang, and Li(2020)}]{ufld}
Qin, Z.; Wang, H.; and Li, X. 2020.
\newblock Ultra fast structure-aware deep lane detection.
\newblock In \emph{Computer Vision--ECCV 2020: 16th European Conference, Glasgow, UK, August 23--28, 2020, Proceedings, Part XXIV 16}, 276--291.

\bibitem[{Qin, Zhang, and Li(2022{\natexlab{a}})}]{ufldv2}
Qin, Z.; Zhang, P.; and Li, X. 2022{\natexlab{a}}.
\newblock Ultra fast deep lane detection with hybrid anchor driven ordinal classification.
\newblock \emph{IEEE Transactions on Pattern Analysis and Machine Intelligence}.

\bibitem[{Qin, Zhang, and Li(2022{\natexlab{b}})}]{ufldv2_simp}
Qin, Z.; Zhang, P.; and Li, X. 2022{\natexlab{b}}.
\newblock Ultra fast deep lane detection with hybrid anchor driven ordinal classification.
\newblock \emph{TPAMI}.

\bibitem[{Qu et~al.(2021)Qu, Jin, Zhou, Yang, and Zhang}]{folo}
Qu, Z.; Jin, H.; Zhou, Y.; Yang, Z.; and Zhang, W. 2021.
\newblock Focus on local: Detecting lane marker from bottom up via key point.
\newblock In \emph{Proceedings of the IEEE/CVF Conference on Computer Vision and Pattern Recognition}, 14122--14130.

\bibitem[{Ren et~al.(2015)Ren, He, Girshick, and Sun}]{fastercnn}
Ren, S.; He, K.; Girshick, R.; and Sun, J. 2015.
\newblock Faster r-cnn: Towards real-time object detection with region proposal networks.
\newblock \emph{Advances in Neural Information Processing Systems}, 28.

\bibitem[{Su et~al.(2021)Su, Chen, Zhang, Luo, Wei, and Wei}]{sgnet}
Su, J.; Chen, C.; Zhang, K.; Luo, J.; Wei, X.; and Wei, X. 2021.
\newblock Structure guided lane detection.
\newblock \emph{arXiv preprint arXiv:2105.05403}.

\bibitem[{Sun et~al.(2021)Sun, Zhang, Jiang, Kong, Xu, Zhan, Tomizuka, Li, Yuan, Wang et~al.}]{sparsercnn}
Sun, P.; Zhang, R.; Jiang, Y.; Kong, T.; Xu, C.; Zhan, W.; Tomizuka, M.; Li, L.; Yuan, Z.; Wang, C.; et~al. 2021.
\newblock Sparse r-cnn: End-to-end object detection with learnable proposals.
\newblock In \emph{Proceedings of the IEEE/CVF Conference on Computer Vision and Pattern Recognition}, 14454--14463.

\bibitem[{Tabelini et~al.(2021{\natexlab{a}})Tabelini, Berriel, Paixao, Badue, De~Souza, and Oliveira-Santos}]{laneatt}
Tabelini, L.; Berriel, R.; Paixao, T.~M.; Badue, C.; De~Souza, A.~F.; and Oliveira-Santos, T. 2021{\natexlab{a}}.
\newblock Keep your eyes on the lane: Real-time attention-guided lane detection.
\newblock In \emph{Proceedings of the IEEE/CVF Conference on Computer Vision and Pattern Recognition}, 294--302.

\bibitem[{Tabelini et~al.(2021{\natexlab{b}})Tabelini, Berriel, Paixao, Badue, De~Souza, and Oliveira-Santos}]{laneatt_simp}
Tabelini, L.; Berriel, R.; Paixao, T.~M.; Badue, C.; De~Souza, A.~F.; and Oliveira-Santos, T. 2021{\natexlab{b}}.
\newblock Keep your eyes on the lane: Real-time attention-guided lane detection.
\newblock In \emph{CVPR}.

\bibitem[{Tabelini et~al.(2021{\natexlab{c}})Tabelini, Berriel, Paixao, Badue, De~Souza, and Oliveira-Santos}]{polylane}
Tabelini, L.; Berriel, R.; Paixao, T.~M.; Badue, C.; De~Souza, A.~F.; and Oliveira-Santos, T. 2021{\natexlab{c}}.
\newblock Polylanenet: Lane estimation via deep polynomial regression.
\newblock In \emph{2020 25th International Conference on Pattern Recognition (ICPR)}, 6150--6156.

\bibitem[{Tan, Pang, and Le(2020)}]{efficientdet}
Tan, M.; Pang, R.; and Le, Q.~V. 2020.
\newblock Efficientdet: Scalable and efficient object detection.
\newblock In \emph{Proceedings of the IEEE/CVF Conference on Computer Vision and Pattern Recognition}, 10781--10790.

\bibitem[{TuSimple(2017)}]{tusimple}
TuSimple. 2017.
\newblock Tusimple lane detection challenge.
\newblock In \emph{CVPR Workshops}.

\bibitem[{Vaswani et~al.(2017)Vaswani, Shazeer, Parmar, Uszkoreit, Jones, Gomez, Kaiser, and Polosukhin}]{transformer}
Vaswani, A.; Shazeer, N.; Parmar, N.; Uszkoreit, J.; Jones, L.; Gomez, A.~N.; Kaiser, {\L}.; and Polosukhin, I. 2017.
\newblock Attention is all you need.
\newblock \emph{Advances in Neural Information Processing Systems}, 30.

\bibitem[{Wang et~al.(2022)Wang, Ma, Huang, Hui, Wang, Qian, and Zhang}]{ganet}
Wang, J.; Ma, Y.; Huang, S.; Hui, T.; Wang, F.; Qian, C.; and Zhang, T. 2022.
\newblock A keypoint-based global association network for lane detection.
\newblock In \emph{Proceedings of the IEEE/CVF Conference on Computer Vision and Pattern Recognition}, 1392--1401.

\bibitem[{Wu, Chang, and Lin(2014)}]{markerextraction}
Wu, P.-C.; Chang, C.-Y.; and Lin, C.~H. 2014.
\newblock Lane-mark extraction for automobiles under complex conditions.
\newblock \emph{Pattern Recognition}, 47(8): 2756--2767.

\bibitem[{Xu et~al.(2020)Xu, Wang, Cai, Zhang, Liang, and Li}]{curvelane}
Xu, H.; Wang, S.; Cai, X.; Zhang, W.; Liang, X.; and Li, Z. 2020.
\newblock Curvelane-nas: Unifying lane-sensitive architecture search and adaptive point blending.
\newblock In \emph{Computer Vision--ECCV 2020: 16th European Conference, Glasgow, UK, August 23--28, 2020, Proceedings, Part XV 16}, 689--704.

\bibitem[{Xu et~al.(2022{\natexlab{a}})Xu, Cai, Zhao, Zhang, Xu, Fu, and Xue}]{rclane}
Xu, S.; Cai, X.; Zhao, B.; Zhang, L.; Xu, H.; Fu, Y.; and Xue, X. 2022{\natexlab{a}}.
\newblock RCLane: Relay Chain Prediction for Lane Detection.
\newblock In \emph{Computer Vision--ECCV 2022: 17th European Conference, Tel Aviv, Israel, October 23--27, 2022, Proceedings, Part XXXVIII}, 461--477.

\bibitem[{Xu et~al.(2022{\natexlab{b}})Xu, Cai, Zhao, Zhang, Xu, Fu, and Xue}]{rclane_simp}
Xu, S.; Cai, X.; Zhao, B.; Zhang, L.; Xu, H.; Fu, Y.; and Xue, X. 2022{\natexlab{b}}.
\newblock RCLane: Relay Chain Prediction for Lane Detection.
\newblock In \emph{ECCV}.

\bibitem[{Yang, Huang, and Wang(2022)}]{querydet}
Yang, C.; Huang, Z.; and Wang, N. 2022.
\newblock Querydet: Cascaded sparse query for accelerating high-resolution small object detection.
\newblock In \emph{Proceedings of the IEEE/CVF Conference on Computer Vision and Pattern Recognition}, 13668--13677.

\bibitem[{Yu et~al.(2021)Yu, Xia, Bai, Lu, Yuille, and Shen}]{ggvit}
Yu, Q.; Xia, Y.; Bai, Y.; Lu, Y.; Yuille, A.~L.; and Shen, W. 2021.
\newblock Glance-and-gaze vision transformer.
\newblock \emph{Advances in Neural Information Processing Systems}, 34: 12992--13003.

\bibitem[{Zheng et~al.(2022{\natexlab{a}})Zheng, Huang, Liu, Tang, Yang, Cai, and He}]{clrnet}
Zheng, T.; Huang, Y.; Liu, Y.; Tang, W.; Yang, Z.; Cai, D.; and He, X. 2022{\natexlab{a}}.
\newblock Clrnet: Cross layer refinement network for lane detection.
\newblock In \emph{Proceedings of the IEEE/CVF Conference on Computer Vision and Pattern Recognition}, 898--907.

\bibitem[{Zheng et~al.(2022{\natexlab{b}})Zheng, Huang, Liu, Tang, Yang, Cai, and He}]{clrnet_simp}
Zheng, T.; Huang, Y.; Liu, Y.; Tang, W.; Yang, Z.; Cai, D.; and He, X. 2022{\natexlab{b}}.
\newblock Clrnet: Cross layer refinement network for lane detection.
\newblock In \emph{CVPR}.

\end{thebibliography}

\end{document}